\documentclass{article}

\usepackage{graphicx} 
\usepackage{subfigure} 
\usepackage{amssymb, amsmath, amsthm}

\usepackage{natbib}

\usepackage{algorithm}
\usepackage{algorithmic}
\usepackage{url}

\usepackage[accepted]{icml2011}

\newcommand{\reals}{\mathbb{R}}

\newtheorem{ass}{Assumption}[section]

\newtheorem{thm}{Theorem}[section]

\newtheorem{lem}[thm]{Lemma}

\newcommand{\figref}[1]{Fig.~\ref{#1}}

\newcommand{\secref}[1]{Sec.~\ref{#1}}

\newcommand{\thmref}[1]{Theorem~\ref{#1}}
\newcommand{\assref}[1]{Assumption~\ref{#1}}

\newcommand{\lemref}[1]{Lemma~\ref{#1}}

\newcommand{\algref}[1]{Alg.~\ref{#1}}

\newcommand{\Ex}[1]{\mathbf{E} \left[ #1 \right] }
\newcommand{\Expect}[2]{\mathbf{E}_{#1} \left[ #2 \right] }

\newcommand{\A}{\mathbf{A}}
\newcommand{\y}{\mathbf{y}}
\newcommand{\x}{\mathbf{x}}
\newcommand{\p}{{\tt P}} 
\newcommand{\Pt}{\mathcal{P}_t} 
\newcommand{\Y}{\mathcal{Y}} 
\newcommand{\ellone}{L_1}

\long\def\symbolfootnote[#1]#2{\begingroup%
\def\thefootnote{\fnsymbol{footnote}}\footnote[#1]{#2}\endgroup}

\newenvironment{changemargin}[2]{%
\begin{list}{}{%
\setlength{\topsep}{0pt}%
\setlength{\leftmargin}{#1}%
\setlength{\rightmargin}{#2}%
\setlength{\listparindent}{\parindent}%
\setlength{\itemindent}{\parindent}%
\setlength{\parsep}{\parskip}%
}%
\item[]}{\end{list}}

\icmltitlerunning{Parallel Coordinate Descent for $\ellone$-Regularized Loss Minimization}

\begin{document} 

\twocolumn[
\icmltitle{Parallel Coordinate Descent for $\ellone$-Regularized Loss Minimization}


\icmlauthor{Joseph K. Bradley $\dagger$}{jkbradle@cs.cmu.edu}
\icmlauthor{Aapo Kyrola $\dagger$}{akyrola@cs.cmu.edu}
\icmlauthor{Danny Bickson}{bickson@cs.cmu.edu}
\icmlauthor{Carlos Guestrin}{guestrin@cs.cmu.edu}
\icmladdress{Carnegie Mellon University,
  5000 Forbes Ave., Pittsburgh, PA 15213 USA}

\icmlkeywords{Lasso, sparse logistic regression, parallel optimization, coordinate descent}

\vskip 0.3in
]

\begin{abstract}
We propose Shotgun, a parallel coordinate descent algorithm for minimizing $\ellone$-regularized losses.
Though coordinate descent seems inherently sequential, we prove convergence bounds for Shotgun which predict linear speedups,
up to a problem-dependent limit.
We present a comprehensive empirical study of Shotgun for Lasso and sparse logistic regression.
Our theoretical predictions on the potential for parallelism closely match behavior on real data.
Shotgun outperforms other published solvers on a range of large problems, proving to be one of the most scalable algorithms for $\ellone$.
\end{abstract} 

\section{Introduction}

Many applications use $\ellone$-regularized models such as the Lasso
\cite{tibshirani96regression} and sparse logistic regression \cite{Ng:ICML04}.
$\ellone$ regularization biases learning towards sparse solutions,
and it is especially useful for \emph{high-dimensional} problems with large numbers of features.
For example, in logistic regression, it allows sample complexity to scale logarithmically w.r.t. the number of irrelevant features \cite{Ng:ICML04}.

Much effort has been put into developing optimization algorithms
for $\ellone$ models. These algorithms range from coordinate minimization
\cite{Shooting} and stochastic gradient \cite{Shalev-Shwartz+Tewari:ICML09}
to more complex interior point methods \cite{L1Boyd}.

Coordinate descent, which we call \textit{Shooting} after Fu \yrcite{Shooting},
is a simple but very effective algorithm which updates one coordinate
per iteration.
It often requires no tuning of parameters, unlike, e.g., stochastic gradient.
As we discuss in \secref{sec:setup},
theory \cite{Shalev-Shwartz+Tewari:ICML09} and extensive empirical results
\cite{LibLinear} have shown that variants of Shooting are particularly competitive for high-dimensional data.

The need for scalable optimization is growing as more applications
use high-dimensional data, but processor core speeds have stopped increasing in
recent years. Instead, computers come with more cores, and the new challenge is utilizing them efficiently.
Yet despite the many sequential optimization algorithms for
$\ellone$-regularized losses, few parallel algorithms exist. 

Some algorithms, such as interior point methods,
can benefit from parallel matrix-vector operations.
However, we found empirically that such algorithms were often outperformed by Shooting.



Recent work analyzes parallel stochastic gradient descent for
multicore \cite{Langford+al:NIPS09} and distributed settings
\cite{Mann+al:NIPS09, SmolaGD}.
These methods parallelize over samples.
In applications using $\ellone$ regularization, though, there are often many
more features than samples,
so parallelizing over samples may be of limited utility.
\symbolfootnote[0]{$\dagger$ These authors contributed equally to this work.}

We therefore take an orthogonal approach and parallelize over features,
with a remarkable result: we can parallelize coordinate descent---an
algorithm which seems inherently sequential---for $\ellone$-regularized losses.
In \secref{sec:parallelCD}, we propose \textit{Shotgun}, a simple
multicore algorithm which makes $\p$ coordinate updates in parallel.
We prove strong convergence bounds for Shotgun which predict speedups over
Shooting which are linear in $\p$, up to a problem-dependent maximum $\p^*$. 
Moreover, our theory provides an estimate for this ideal $\p^*$ which may be
easily computed from the data. 

Parallel coordinate descent was also considered by Tsitsiklis et al.
\yrcite{Tsitsiklis+al:IEEE86},
but for differentiable objectives in the asynchronous setting.
They give a very general analysis, proving asymptotic convergence but not
convergence rates.  We are able to prove rates and theoretical speedups
for our class of objectives.

In \secref{sec:exp:results}, we compare multicore Shotgun with five
state-of-the-art algorithms on 35 real and synthetic datasets. The results
show that in large problems Shotgun outperforms the other algorithms.
Our experiments also validate the theoretical predictions by showing that
Shotgun requires only about $1/\p$ as many iterations as Shooting.
We measure the parallel speedup in running time and analyze the
limitations imposed by the multicore hardware.


\section{$\ellone$-Regularized Loss Minimization \label{sec:setup} }

We consider optimization problems of the form
\small
\begin{equation} \label{eqn:objective:general}
  \min \limits_{\x \in \reals^d} F(\x)
  = \sum_{i=1}^n L(\mathbf{a}_{i}^T \x, y_i) + \lambda \lVert \x \rVert_1 \,,
\end{equation} \normalsize
where $L(\cdot)$ is a non-negative convex loss. Each of $n$ \emph{samples} has a
\emph{feature vector} $\mathbf{a}_i \in \reals^d$ and observation $y_i$
(where $\y \in \Y^n$).
$\x \in \reals^d$ is an \emph{unknown} vector of \emph{weights} for features.
$\lambda \ge 0$ is a regularization parameter.
Let $\A \in \reals^{n \times d}$ be the design matrix, whose $i^{\textrm{th}}$
row is $\mathbf{a}_i$. Assume w.l.o.g. that columns of $\A$ are
normalized s.t. $\mathop{diag}(\A^T\A) = \mathbf{1}$.\footnote{Normalizing $\A$ does not change the objective if a separate, normalized $\lambda_j$ is used for each $x_j$.}

An instance of \eqref{eqn:objective:general} is the Lasso
\cite{tibshirani96regression} (in penalty form), for which $\Y \equiv \reals$
and
\small
\begin{equation} \label{eqn:objective:lasso}
F(\x) =
\tfrac{1}{2} \lVert \A\x - \y \rVert^2_2 + \lambda \lVert \x \rVert_1 \,,
\end{equation} \normalsize
as well as sparse logistic regression \cite{Ng:ICML04},
for which $\Y \equiv \{-1,+1\}$ and
\small
\begin{equation} \label{eqn:objective:logreg}
F(\x) =
\sum_{i=1}^n \log\left(1 + \exp\left(-y_i\mathbf{a}_i^{T}\x\right)\right) + \lambda \lVert \x \rVert_1 \,.
\end{equation} \normalsize
For analysis, we follow Shalev-Shwartz and Tewari
\yrcite{Shalev-Shwartz+Tewari:ICML09} and transform
\eqref{eqn:objective:general} into an equivalent problem with a
twice-differentiable regularizer.
We let $\hat\x \in \reals^{2d}_+$, use \textit{duplicated features}
$\hat{\mathbf{a}}_i = [\mathbf{a}_i; -\mathbf{a}_i] \in \reals^{2d}$, and solve
\small
\begin{equation} \label{eqn:objective:general2}
\min_{\hat{\x} \in \reals^{2d}_+}
\sum_{i=1}^n L(\hat{\mathbf{a}}_i^T \hat{\x}, y_i) + \lambda \sum_{j=1}^{2d} \hat{x}_j \,.
\end{equation}\normalsize
If $\mathbf{\hat{x}} \in \reals^{2d}_+$ minimizes
\eqref{eqn:objective:general2},
then $\x: \, x_i = \hat{x}_{d+i} - \hat{x}_{i}$ minimizes
\eqref{eqn:objective:general}.
Though our analysis uses duplicate features, they are not needed for an
implementation.

\subsection{Sequential Coordinate Descent \label{sec:sequentialCD} }

\begin{algorithm}[t]
  \caption{Shooting: Sequential SCD}
  \label{alg:shooting}
  \begin{algorithmic}
    \STATE Set $\x = \mathbf{0} \in \reals^{2d}_+$.
    \WHILE{not converged}
    \STATE Choose $j \in \{1,\ldots,2d\}$ uniformly at random.
    \STATE Set $\delta x_j \longleftarrow \max\{-x_j, -(\nabla F(\x))_j/\beta\}$.
    \STATE Update $x_j \longleftarrow x_j + \delta x_j$.
    \ENDWHILE
  \end{algorithmic}
\end{algorithm}

Shalev-Shwartz and Tewari \yrcite{Shalev-Shwartz+Tewari:ICML09} analyze
Stochastic Coordinate Descent (SCD), a stochastic version of Shooting for
solving \eqref{eqn:objective:general}. SCD (\algref{alg:shooting})
randomly chooses one weight $x_j$ to update per iteration.
It computes the update $x_j \leftarrow x_j + \delta x_j$ via
\small
\begin{equation} \label{eqn:shooting:update}
  \delta x_j = \max\{ -x_j, -(\nabla F(\x))_j / \beta \} \,,
\end{equation}\normalsize
where $\beta>0$ is a loss-dependent constant.

To our knowledge, Shalev-Shwartz and Tewari
\yrcite{Shalev-Shwartz+Tewari:ICML09} provide the best known
convergence bounds for SCD.
Their analysis requires a uniform upper bound on the change in the loss $F(\x)$
from updating a single weight:
\begin{ass} \label{ass:uniform:upper:bound}
  Let $F(\x): \reals^{2d}_+ \longrightarrow \reals$ be a convex function.
  Assume there exists $\beta > 0$ s.t., for all $\x$ and single-weight updates
  $\delta x_j$, we have:
  \begin{equation*}
    F(\x + (\delta x_j) \mathbf{e}^j)
    \le F(\x) + \delta x_j (\nabla F(\x))_j + \tfrac{\beta (\delta x_j)^2}{2} \,,
  \end{equation*}
\end{ass}
where $\mathbf{e}^j$ is a unit vector with $1$ in its $j^{\mathrm{th}}$ entry.
For the losses in \eqref{eqn:objective:lasso} and
\eqref{eqn:objective:logreg}, Taylor expansions give
\begin{equation} \label{eqn:betas}
  \beta = 1 \, \mbox{(squared loss) and} \,\, \beta = \tfrac{1}{4} \, \mbox{(logistic loss).}
\end{equation}
Using this bound, they prove the following theorem.

\begin{thm} \cite{Shalev-Shwartz+Tewari:ICML09} \label{thm:shooting}
  Let $\x^*$ minimize \eqref{eqn:objective:general2} and $\x^{(T)}$ be the
  output of \algref{alg:shooting} after $T$ iterations.
  If $F(\x)$ satisfies \assref{ass:uniform:upper:bound}, then
  \small\begin{equation}
    \Ex{F(\x^{(T)})} - F(\x^*)
    \le \frac{d(\beta \lVert \x^* \rVert_2^2 + 2F(\x^{(0)}))}{T+1} \,,
  \end{equation}\normalsize
  where $\mathbf{E}[\cdot]$ is w.r.t. the random choices of weights $j$.
\end{thm}

As Shalev-Shwartz and Tewari \yrcite{Shalev-Shwartz+Tewari:ICML09} argue,
\thmref{thm:shooting} indicates that SCD scales well in the dimensionality
$d$ of the data. 
For example, it achieves better runtime bounds w.r.t. $d$ than
stochastic gradient methods such as SMIDAS \cite{Shalev-Shwartz+Tewari:ICML09}
and truncated gradient \cite{Langford+al:truncgrad}.



\section{Parallel Coordinate Descent \label{sec:parallelCD} }

As the dimensionality $d$ or sample size $n$ increase,
even fast sequential algorithms become expensive.
To scale to larger problems, we turn to parallel computation.
In this section, we present our main theoretical contribution:
we show coordinate descent can be parallelized by proving strong convergence
bounds.

\begin{algorithm}[t]
  \caption{Shotgun: Parallel SCD}
  \label{alg:shotgun}
  \begin{algorithmic}
     \STATE Choose number of parallel updates $\p \ge 1$.
    \STATE Set $\x = \mathbf{0} \in \reals^{2d}_+$
    \WHILE{not converged}
    \INPARALLEL{on $\p$ processors}
    \STATE Choose $j \in \{1,\ldots,2d\}$ uniformly at random.
    \STATE Set $\delta x_j \longleftarrow \max\{-x_j, -(\nabla F(\x))_j/\beta\}$.
    \STATE Update $x_j \longleftarrow x_j + \delta x_j$.
    \ENDINPARALLEL
    \ENDWHILE
  \end{algorithmic}
\end{algorithm}

We parallelize stochastic Shooting and call our algorithm
Shotgun (\algref{alg:shotgun}). Shotgun initially chooses $\p$,
the number of weights to update in parallel.
On each iteration, it chooses $\p$ weights independently and uniformly at random
from $\{1,\ldots,2d\}$; these form a multiset $\Pt$.
It updates each $x_{i_j}: \, i_j \in \Pt,$ in parallel using the same update
as Shooting \eqref{eqn:shooting:update}.
Let $\Delta \x$ be the collective update to $\x$, i.e.,
$(\Delta \x)_k = \sum_{i_j \in \Pt : \, k = i_j} \delta x_{i_j}$.

\begin{figure}[t]
  \centering
  \begin{tabular}{cc}
    \includegraphics[scale=.28]{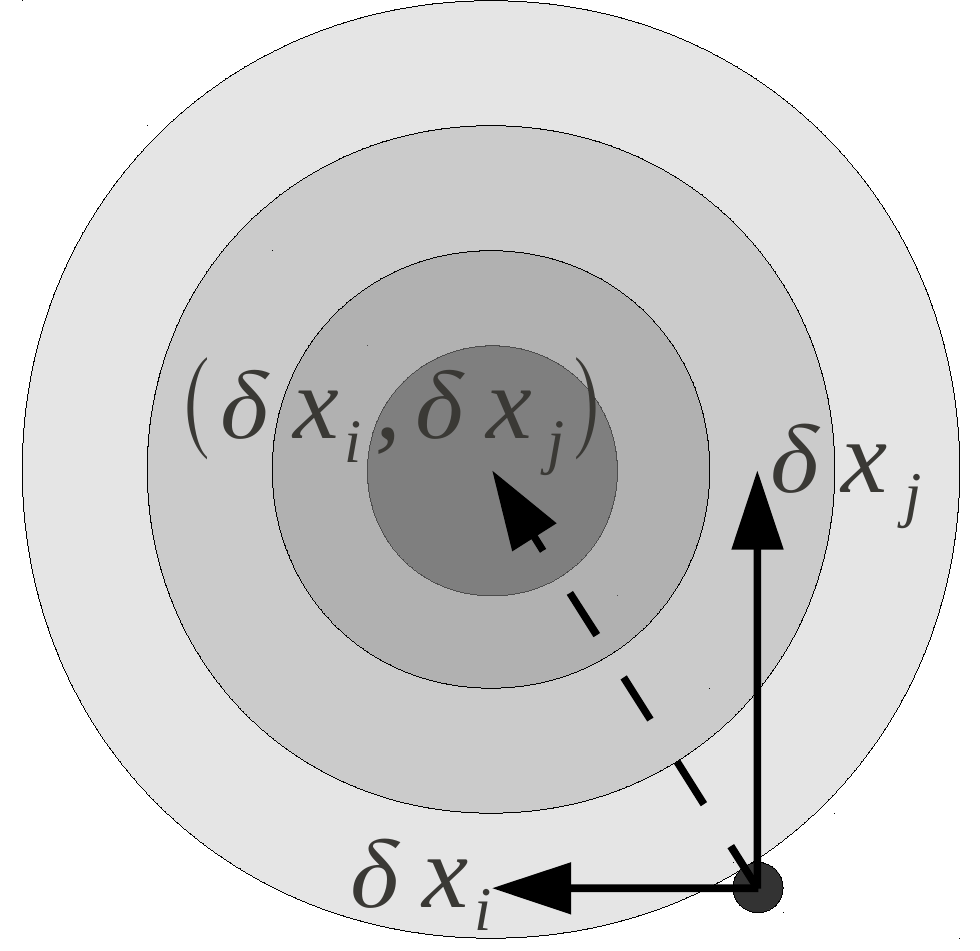} &
    \includegraphics[scale=.28]{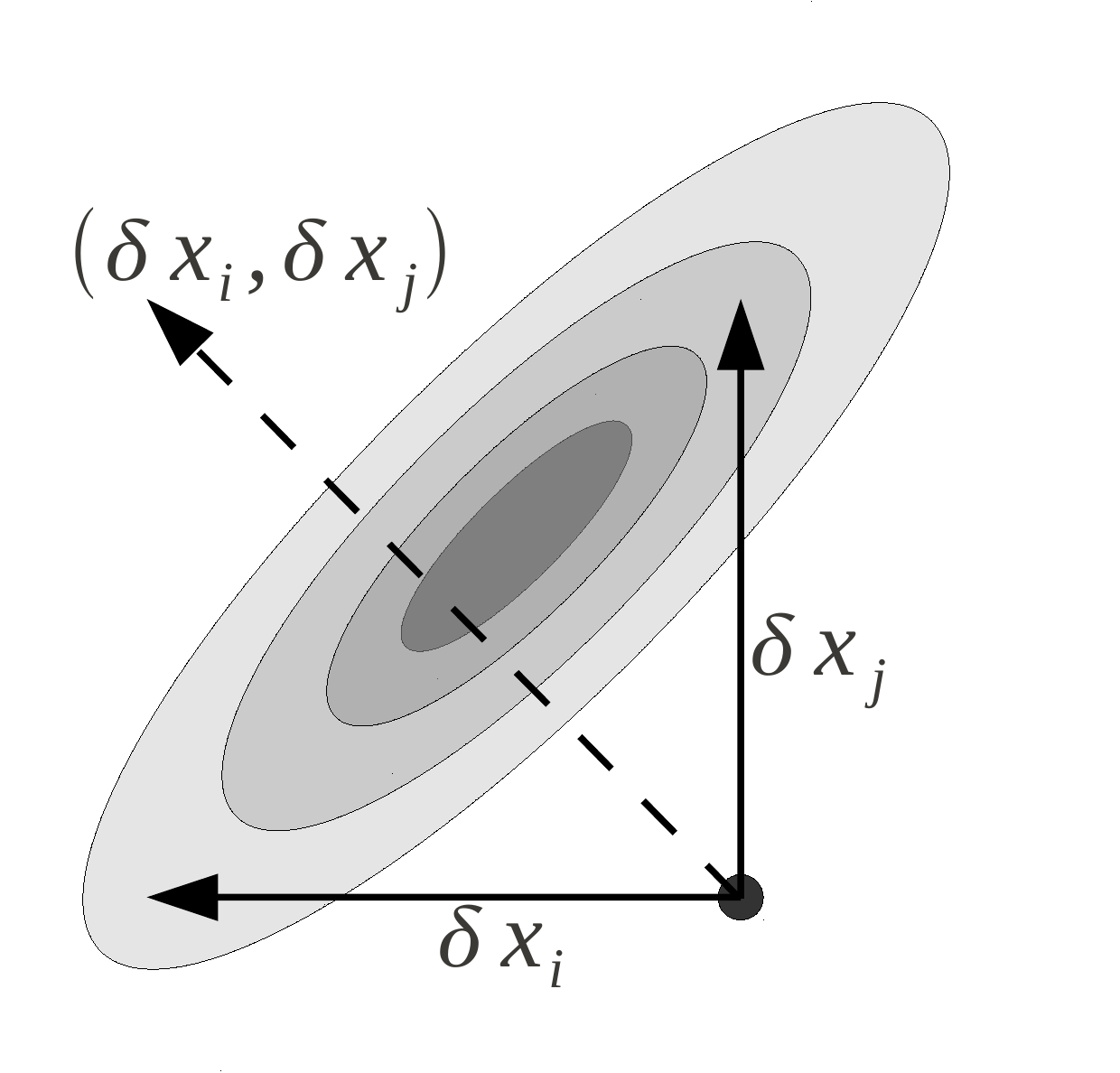}
  \end{tabular}
  \caption{Intuition for parallel coordinate descent.
    Contour plots of two objectives, with darker meaning better.
    Left: Features are uncorrelated; parallel updates are useful.
    Right: Features are correlated; parallel updates conflict.
    \label{fig:shotgun:intuition} }
\end{figure}

Intuitively, parallel updates might increase the risk of divergence.
In \figref{fig:shotgun:intuition}, in the left subplot,
parallel updates speed up convergence since features are uncorrelated;
in the right subplot, parallel updates of correlated features risk
increasing the objective. We can avoid divergence by imposing a step
size, but our experiments showed that approach to be impractical.\footnote{A step size of $\tfrac{1}{\p}$ ensures convergence since $F$ is convex in $\x$, but it results in very small steps and long runtimes.}

We formalize this intuition for the Lasso in \thmref{thm:shotgun:intuition}.
We can separate a sequential progress term (summing the improvement from
separate updates) from a term measuring interference between parallel updates.
If $\A^T\A$ were normalized and centered to be a covariance matrix,
the elements in the interference term's sum would be non-zero only for
correlated variables, matching our intuition from
\figref{fig:shotgun:intuition}. Harmful interference could occur when, e.g.,
$\delta x_i, \delta x_j > 0$ and features $i,j$ were positively correlated.

\begin{thm} \label{thm:shotgun:intuition}
  Fix $\x$. If $\Delta \x$ is the collective update to $\x$
  in one iteration of \algref{alg:shotgun} for the Lasso, then
  \begin{eqnarray}
    \lefteqn{ F(\x + \Delta \x) - F(\x) } \nonumber \\
    && \le 
    \underbrace{-\tfrac{1}{2} \sum_{i_j \in \Pt} (\delta x_{i_j})^2}_\textrm{sequential progress}
    + \underbrace{\tfrac{1}{2} \sum_{\substack{i_j,i_k \in \Pt, \\ j \ne k}} (\A^T\A)_{i_j,i_k} \delta x_{i_j} \delta x_{i_k}}_\textrm{interference}. \nonumber
  \end{eqnarray}
\end{thm}
\textbf{Proof Sketch}\footnote{We include detailed proofs of all theorems and lemmas in the supplementary material.}:
Write the Taylor expansion of $F$ around $\x$.
Bound the first-order term using \eqref{eqn:shooting:update}. $\blacksquare$ \\
In the next section, we show that this intuition holds for the more general
optimization problem in \eqref{eqn:objective:general}.

\subsection{Shotgun Convergence Analysis \label{sec:shotgun:analysis} }

In this section, we present our convergence result for Shotgun.
The result provides a problem-specific measure of the potential for
parallelization: the spectral radius $\rho$ of $\A^T\A$ (i.e., the maximum of
the magnitudes of eigenvalues of $\A^T\A$). Moreover, this measure is
prescriptive: $\rho$ may be estimated via, e.g., power iteration\footnote{For our datasets, power iteration gave reasonable estimates within a small fraction of the total runtime.}
\cite{strang:88}, and it provides a plug-in estimate of the ideal number
of parallel updates.

We begin by generalizing \assref{ass:uniform:upper:bound} to our parallel
setting. The scalars $\beta$ for Lasso and logistic regression remain the same
as in \eqref{eqn:betas}.

\begin{ass} \label{ass:uniform:upper:bound:parallel}
  Let $F(\x): \reals^{2d}_+ \longrightarrow \reals$ be a convex function.
  Assume that there exists $\beta > 0$ such that,
  for all $\x$ and parallel updates $\Delta \x$, we have
  \small\begin{equation*}
    F(\x + \Delta \x)
    \le F(\x) + \Delta \x^T \nabla F(\x) + \tfrac{\beta}{2} \Delta \x^T\A^T\A\Delta \x \,.
  \end{equation*}\normalsize
\end{ass}

We now state our main result, generalizing the convergence bound
in \thmref{thm:shooting} to the Shotgun algorithm.

\begin{thm} \label{thm:shotgun}
  Let $\x^*$ minimize \eqref{eqn:objective:general2} and $\x^{(T)}$ be the
  output of \algref{alg:shotgun} after $T$ iterations with $\p$ parallel updates/iteration. Let $\rho$ be the spectral radius of $\A^T\A$.
  If $F(\x)$ satisfies \assref{ass:uniform:upper:bound:parallel} and $\p < \tfrac{2d}{\rho} + 1$, then
  \begin{equation*}
    \Ex{F(\x^{(T)})} - F(\x^*)
    \le \frac{d\left(\beta \lVert \x^* \rVert_2^2 + 2F(\x^{(0)})\right)}{(T+1)\p} \,,
  \end{equation*}
  where the expectation is w.r.t. the random choices of weights to update.
  Choosing a maximal $\p^* \approx \tfrac{2d}{\rho}$ gives
  \begin{equation*}
    \Ex{F(\x^{(T)})} - F(\x^*)
    \lesssim \frac{\rho \left(\tfrac{\beta}{2} \lVert \x^* \rVert_2^2 + F(\x^{(0)}) \right)}{T+1} \,.
  \end{equation*}
\end{thm}

Without duplicated features, \thmref{thm:shotgun} predicts that we can do up to
$\p < \tfrac{d}{\rho}+1$ parallel updates and achieve speedups linear in $\p$.
We denote the predicted maximum $\p$ as $\p^* \equiv ceiling(\tfrac{d}{\rho})$.
For an ideal problem with uncorrelated features, $\rho = 1$,
so we could do up to $\p^* = d$ parallel updates.
For a pathological problem with exactly correlated features, $\rho = d$,
so our theorem tells us that we could not do parallel updates.
With $\p=1$, we recover the result for Shooting in \thmref{thm:shooting}.

To prove \thmref{thm:shotgun}, we first bound the negative impact of
interference between parallel updates.

\begin{lem} \label{lem:shotgun}
  Fix $\x$.
  Under the assumptions and definitions from \thmref{thm:shotgun},
  if $\Delta \x$ is the collective update to $\x$ in
  one iteration of \algref{alg:shotgun}, then
  \begin{equation*}
    \begin{array}{l}
    \Expect{\Pt}{ F(\x + \Delta \x) - F(\x) } \\
    \quad \le
    \p \Expect{j}{ \delta x_j (\nabla F(\x))_j + \tfrac{\beta}{2} \left( 1 - \tfrac{(\p-1)\rho}{2d} \right) (\delta x_j)^2 } \,,
    \end{array}
  \end{equation*}
  where $\mathbf{E}_{\Pt}$ is w.r.t. a random choice of $\Pt$ and
  $\mathbf{E}_j$ is w.r.t. choosing $j \in \{1,\ldots,2d\}$ uniformly at random.
\end{lem}

\textbf{Proof Sketch}:
Take the expectation w.r.t. $\Pt$ of the inequality in
\assref{ass:uniform:upper:bound:parallel}.
\small
\begin{equation}
  \begin{array}{l}
    \Expect{\Pt}{ F(\x + \Delta \x) - F(\x) } \\
    \quad \le \Expect{\Pt}{ \Delta \x^T \nabla F(\x) + \tfrac{\beta}{2} \Delta \x^T\A^T\A\Delta \x }
  \end{array}
\end{equation} \normalsize
Separate the diagonal elements from the second order term,
and rewrite the expectation using our independent choices of $i_j \in \Pt$.
(Here, $\delta x_j$ is the update given by \eqref{eqn:shooting:update},
regardless of whether $j \in \Pt$.)
\small
\begin{equation} \label{eqn:shotgun:proof1}
  \begin{array}{l}
    = \p \Expect{j}{\delta x_j (\nabla F(\x))_j + \tfrac{\beta}{2}(\delta x_j)^2 } \\
      \quad + \tfrac{\beta}{2} \p(\p-1) \Expect{i}{\Expect{j}{\delta x_i (\A^T\A)_{i,j} \delta x_j} }
  \end{array}
\end{equation} \normalsize
Upper bound the double expectation in terms of $\Expect{j}{(\delta x_j)^2}$
by expressing the spectral radius $\rho$ of $\A^T\A$ as
$\rho = \max_{\mathbf{z}:\, \mathbf{z}^T\mathbf{z} =1} \mathbf{z}^T (\A^T\A) \mathbf{z}$.
\small
\begin{equation} \label{eqn:shotgun:proof2}
  \Expect{i}{\Expect{j}{\delta x_i (\A^T\A)_{i,j} \delta x_j} }
  \le \tfrac{\rho}{2d} \Expect{j}{(\delta x_j)^2}
\end{equation} \normalsize
Combine \eqref{eqn:shotgun:proof2} back into \eqref{eqn:shotgun:proof1},
and rearrange terms to get the lemma's result. $\blacksquare$

\textbf{Proof Sketch (\thmref{thm:shotgun})}:
Our proof resembles Shalev-Shwartz and Tewari
\yrcite{Shalev-Shwartz+Tewari:ICML09}'s proof of \thmref{thm:shooting}.
The result from \lemref{lem:shotgun} replaces \assref{ass:uniform:upper:bound}.
One bound requires $\tfrac{(\p-1)\rho}{2d} < 1$.
$\blacksquare$

Our analysis implicitly assumes that parallel updates of the same
weight $x_j$ will not make $x_j$ negative. Proper write-conflict resolution
can ensure this assumption holds and is viable in our multicore setting.

\subsection{Theory vs. Empirical Performance}

We end this section by comparing the predictions of \thmref{thm:shotgun} about
the number of parallel updates $\p$ with empirical performance for Lasso.
We exactly simulated Shotgun as in \algref{alg:shotgun} to eliminate effects
from the practical implementation choices made in \secref{sec:exp:results}.
We tested two single-pixel camera datasets from Duarte et al.
\yrcite{duarte2008single} with very different $\rho$, estimating
$\Expect{P_t}{F(\x^{(T)})}$ by averaging 10 runs of Shotgun.
We used $\lambda = 0.5$ for {\tt Ball64\_singlepixcam} to get $\x^*$ with about
$27\%$ non-zeros; we used $\lambda = 0.05$ for {\tt Mug32\_singlepixcam}
to get about $20\%$ non-zeros.

\begin{figure}[t]
  \begin{tabular}{l|l}
    \includegraphics[scale=.2]{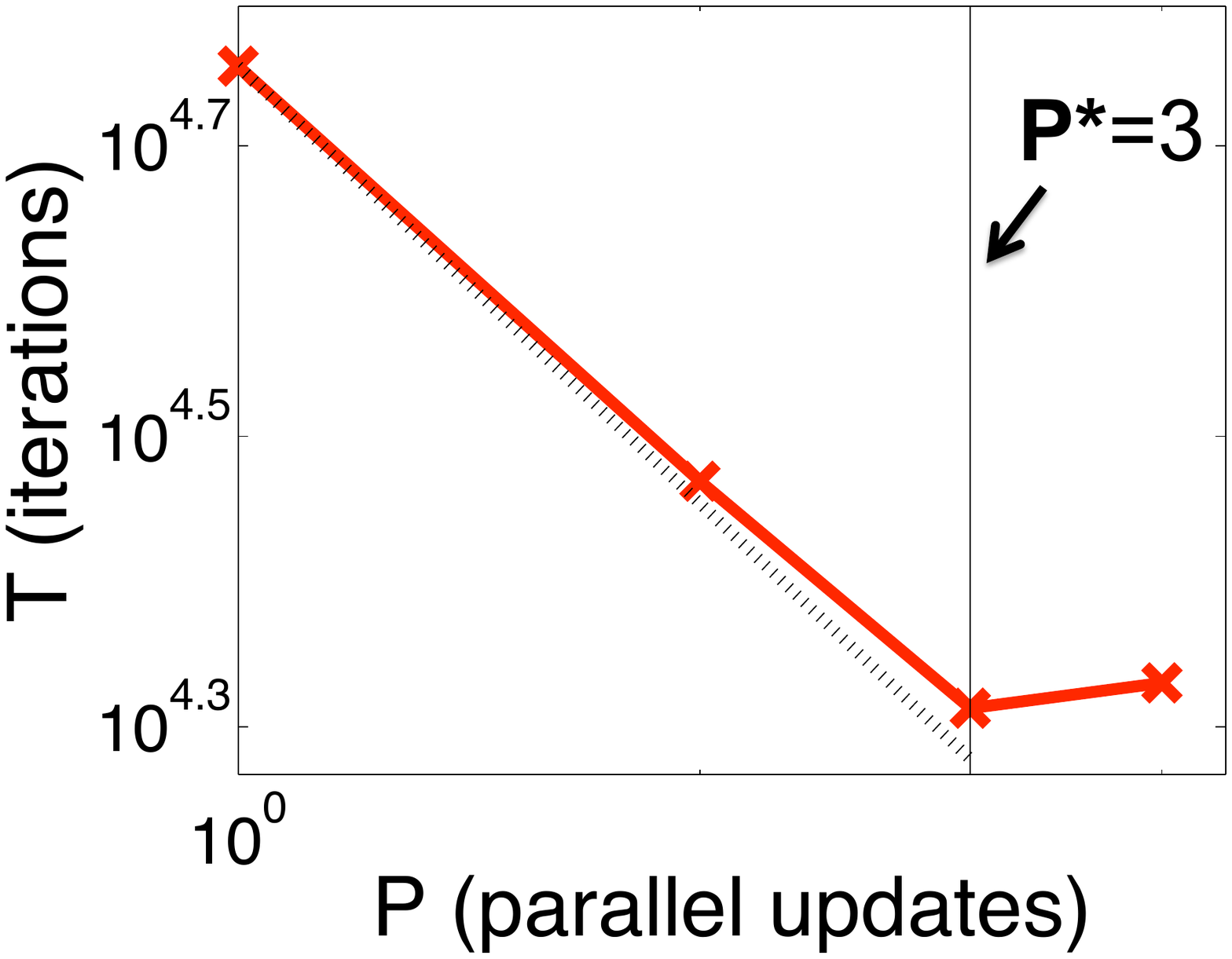} &
    \includegraphics[scale=.2]{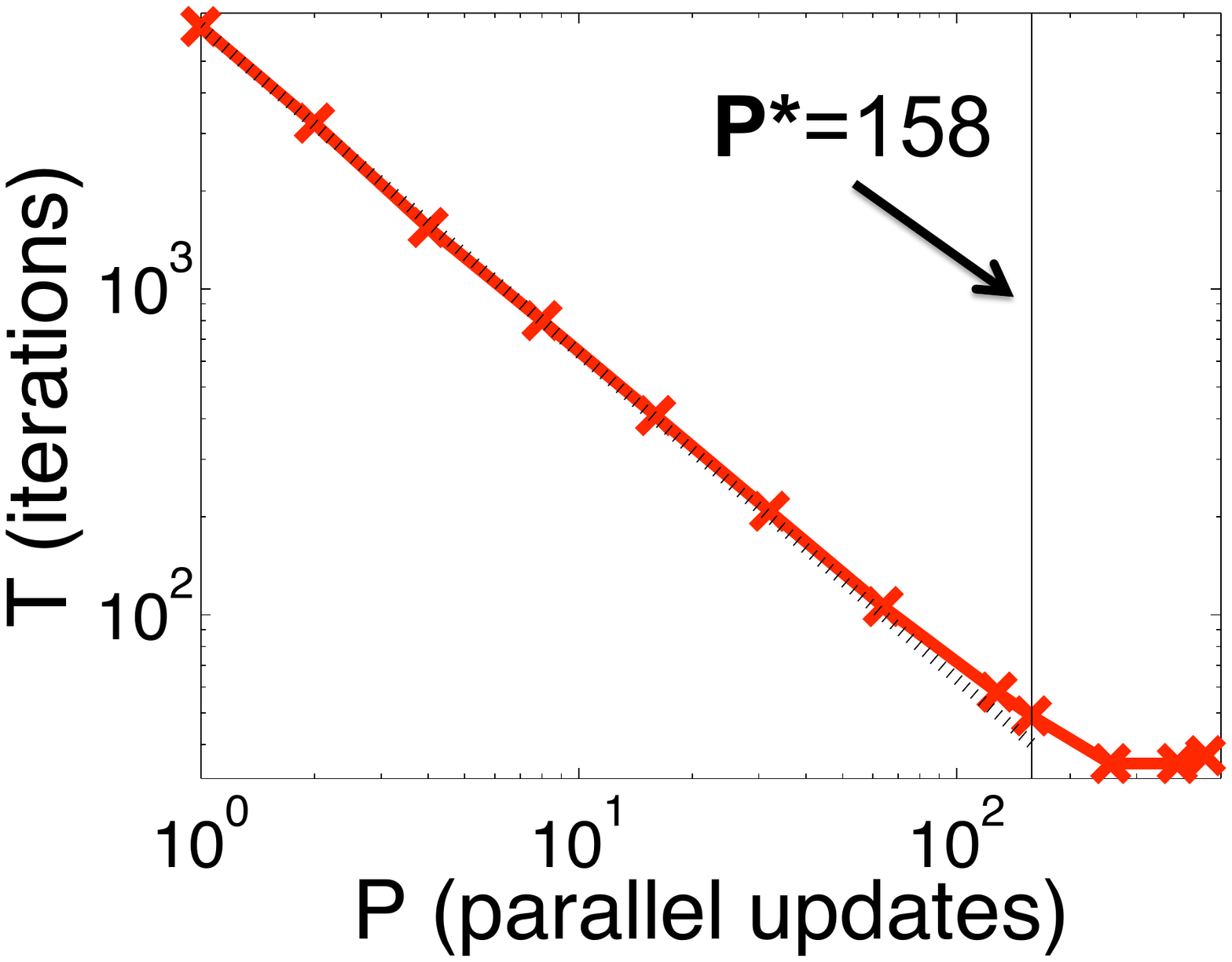} \\
    \small Data: Ball64\_singlepixcam & \small Data: Mug32\_singlepixcam \\
    \small $d = 4096$, $\rho = 2047.8$ & \small $d = 1024$, $\rho = 6.4967$
    \normalsize
  \end{tabular}
  \caption{Theory for Shotgun's $\p$ (\thmref{thm:shotgun}) vs.
    empirical performance for Lasso on two datasets. Y-axis has iterations $T$ until
    $\mathbf{E}_{P_t}[F(\x^{(T)})]$ came within $0.5\%$ of $F(\x^*)$.\\
    Thick red lines trace $T$ for increasing $\p$ (until too large $\p$ caused divergence).
    Vertical lines mark $\p^*$.
    Dotted diagonal lines show optimal (linear) speedups (partly hidden by solid line in right-hand plot).
    \label{fig:shotgun:ptheory} }
\end{figure}

\figref{fig:shotgun:ptheory} plots $\p$ versus the iterations $T$ required for
$\Expect{P_t}{F(\x^{(T)})}$ to come within $0.5\%$ of the optimum $F(\x^*)$.
\thmref{thm:shotgun} predicts that $T$ should decrease as $\tfrac{1}{\p}$
as long as $\p < \p^* \approx \tfrac{d}{\rho} + 1$.
The empirical behavior follows this theory: using the predicted $\p^*$ gives
almost optimal speedups, and speedups are almost linear in $\p$.
As $\p$ exceeds $\p^*$, Shotgun soon diverges.

\figref{fig:shotgun:ptheory} confirms \thmref{thm:shotgun}'s result:
Shooting, a seemingly sequential algorithm, can be parallelized and achieve
near-linear speedups, and the spectral radius of $\A^T\A$ succinctly captures
the potential for parallelism in a problem.
To our knowledge, our convergence results are the first for parallel
coordinate descent for $\ellone$-regularized losses, and they apply to any
convex loss satisfying \assref{ass:uniform:upper:bound:parallel}.
Though \figref{fig:shotgun:ptheory} ignores certain implementation issues,
we show in the next section that Shotgun performs well in practice.

\subsection{Beyond $\ellone$}

Theorems~\ref{thm:shooting} and \ref{thm:shotgun} generalize beyond $\ellone$,
for their main requirements (Assumptions~\ref{ass:uniform:upper:bound},~\ref{ass:uniform:upper:bound:parallel})
apply to a more general class of problems:
$\min F(\x)$ s.t. $\x \ge 0$, where $F(\x)$ is smooth.
We discuss Shooting and Shotgun for sparse regression
since both the method (coordinate descent) and problem (sparse regression)
are arguably most useful for high-dimensional settings.


\section{Experimental Results\label{sec:exp:results}}

We present an extensive study of Shotgun for the Lasso and
sparse logistic regression.  On a wide variety of datasets,
we compare Shotgun with published state-of-the-art solvers.
We also analyze self-speedup in detail in terms of \thmref{thm:shotgun} and hardware issues.

\subsection{Lasso}

We tested Shooting and Shotgun for the Lasso against five published Lasso solvers on 35 datasets.
We summarize the results here; details are in the supplement.

\subsubsection{Implementation: {\tt Shotgun} \label{sec:shotgun:implementation}}

Our implementation made several practical improvements to the basic Shooting and Shotgun algorithms.

Following Friedman et al. \yrcite{friedman2010regularization}, we maintained a
vector $\A\x$ to avoid repeated computation.
We also used their pathwise optimization scheme:
rather than directly solving with the given $\lambda$,
we solved with an exponentially decreasing sequence
$\lambda_1,\lambda_2,\ldots,\lambda$.  The solution $\x$ for $\lambda_k$ is used
to warm-start optimization for $\lambda_{k+1}$.
This scheme can give significant speedups.

Though our analysis is for the synchronous setting, our implementation was
asynchronous because of the high cost of synchronization.
We used atomic compare-and-swap operations for updating the
$\A\x$ vector. 


We used C++ and the CILK++ library \cite{cilkplus} for parallelism.
All tests ran on an AMD processor using up to eight Opteron 8384 cores (2.69 GHz).


%
%



\subsubsection{Other Algorithms}

{\tt L1\_LS} \cite{L1Boyd} is a log-barrier interior point method.
It uses Preconditioned Conjugate Gradient (PCG) to solve Newton steps iteratively and avoid explicitly inverting the Hessian.
The implementation is in Matlab${}^{\textrm{\textregistered}}$, but the expensive step (PCG) uses very efficient native Matlab calls.
In our tests, matrix-vector operations were parallelized on up to 8 cores.

{\tt FPC\_AS} \cite{wen2010fast} uses iterative shrinkage to estimate which
elements of $\x$ should be non-zero, as well as their signs.
This reduces the objective to a smooth, quadratic function which is then minimized.

{\tt GPSR\_BB} \cite{figueiredo2008gradient} is a gradient
projection method which uses line search and termination techniques tailored for the Lasso.

{\tt Hard\_l0} \cite{blumensath2009iterative} uses iterative hard thresholding for compressed sensing.
It sets all but the $s$ largest weights to zero on each iteration.
We set $s$ as the sparsity obtained by {\tt Shooting}.

{\tt SpaRSA} \cite{wright2009sparse} is an accelerated iterative shrinkage/thresholding algorithm which solves a sequence
of quadratic approximations of the objective.

As with {\tt Shotgun}, all of {\tt Shooting}, {\tt FPC\_AS}, {\tt GPSR\_BB}, and {\tt SpaRSA} use pathwise optimization schemes.


We also tested published implementations of the classic algorithms {\tt GLMNET} \cite{friedman2010regularization} and {\tt LARS} \cite{efron2004least}.
Since we were unable to get them to run on our larger datasets, we exclude their results.

\subsubsection{Results}

\begin{figure*}[t]
  \begin{tabular}{cccc}
    \includegraphics[width=0.23\textwidth]{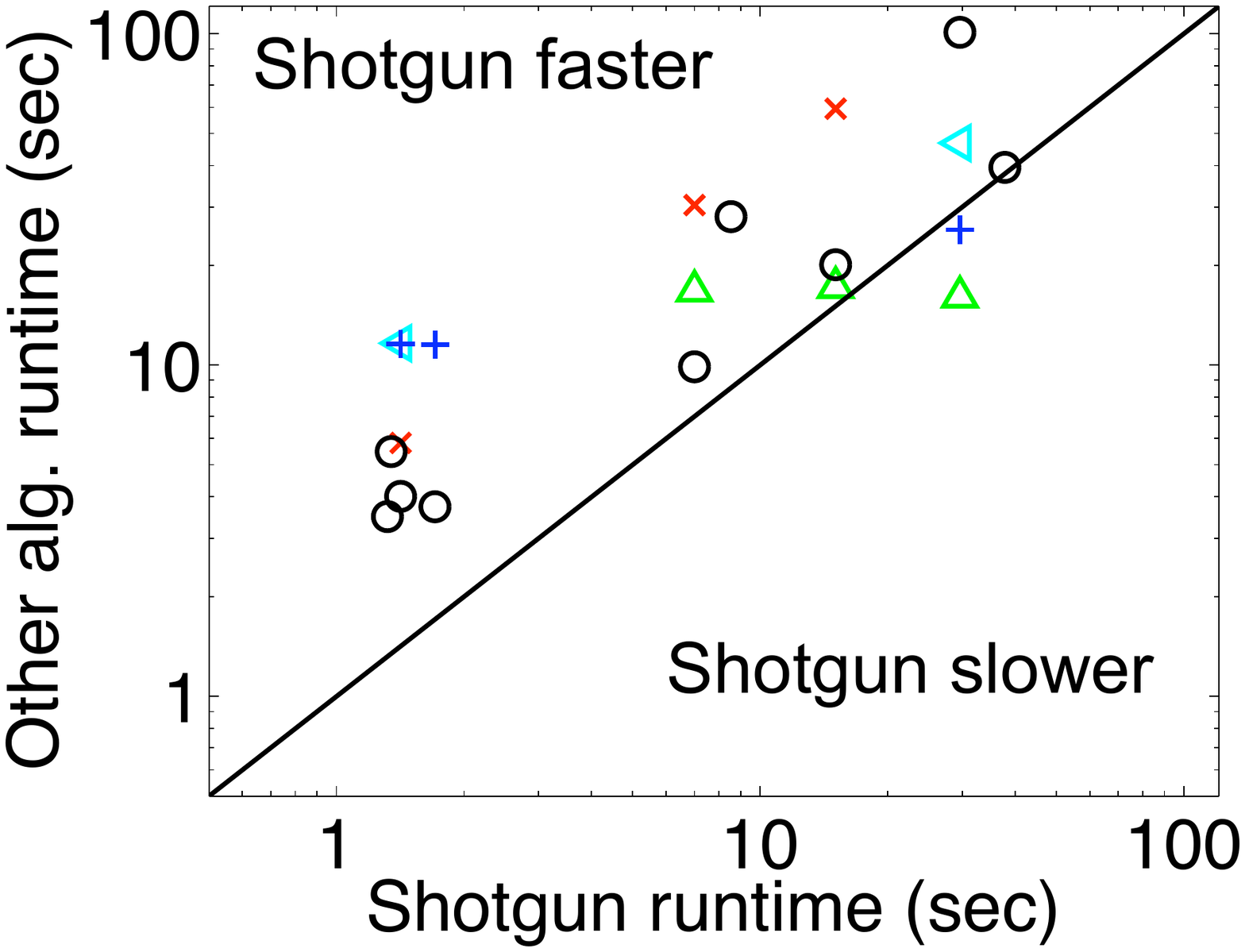} &  \includegraphics[width=0.23\textwidth]{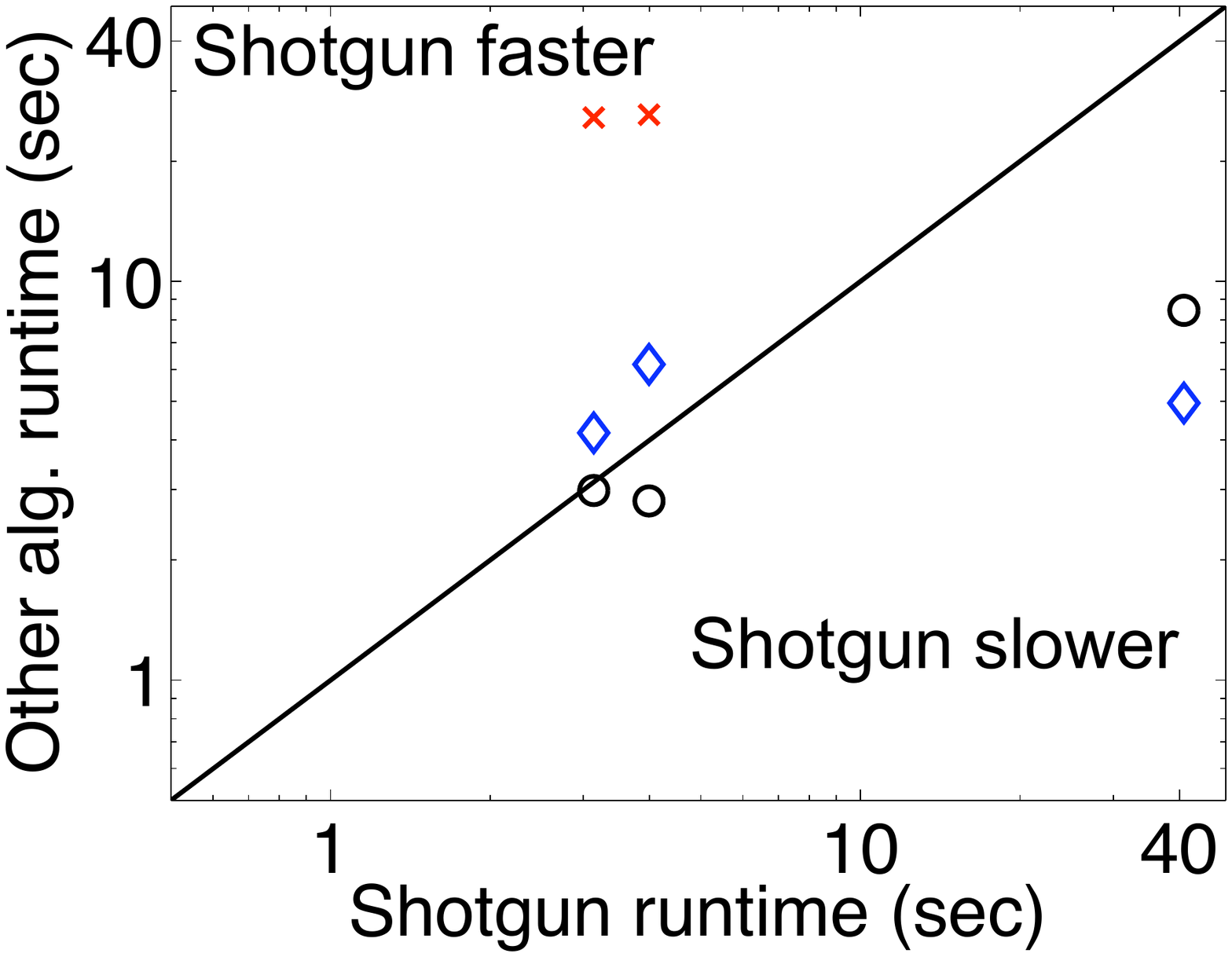} &
    \includegraphics[width=0.23\textwidth]{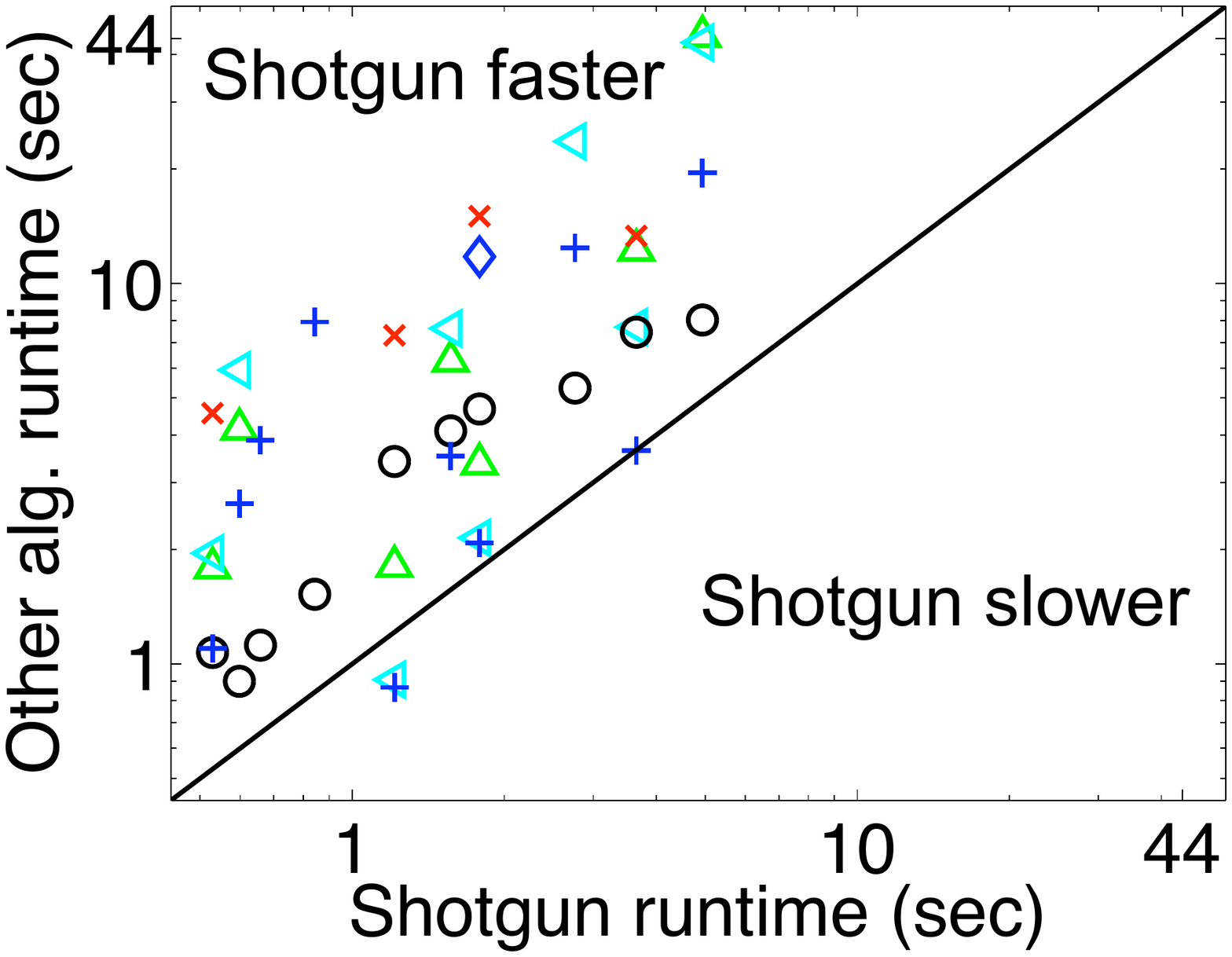}&  \includegraphics[width=0.24\textwidth]{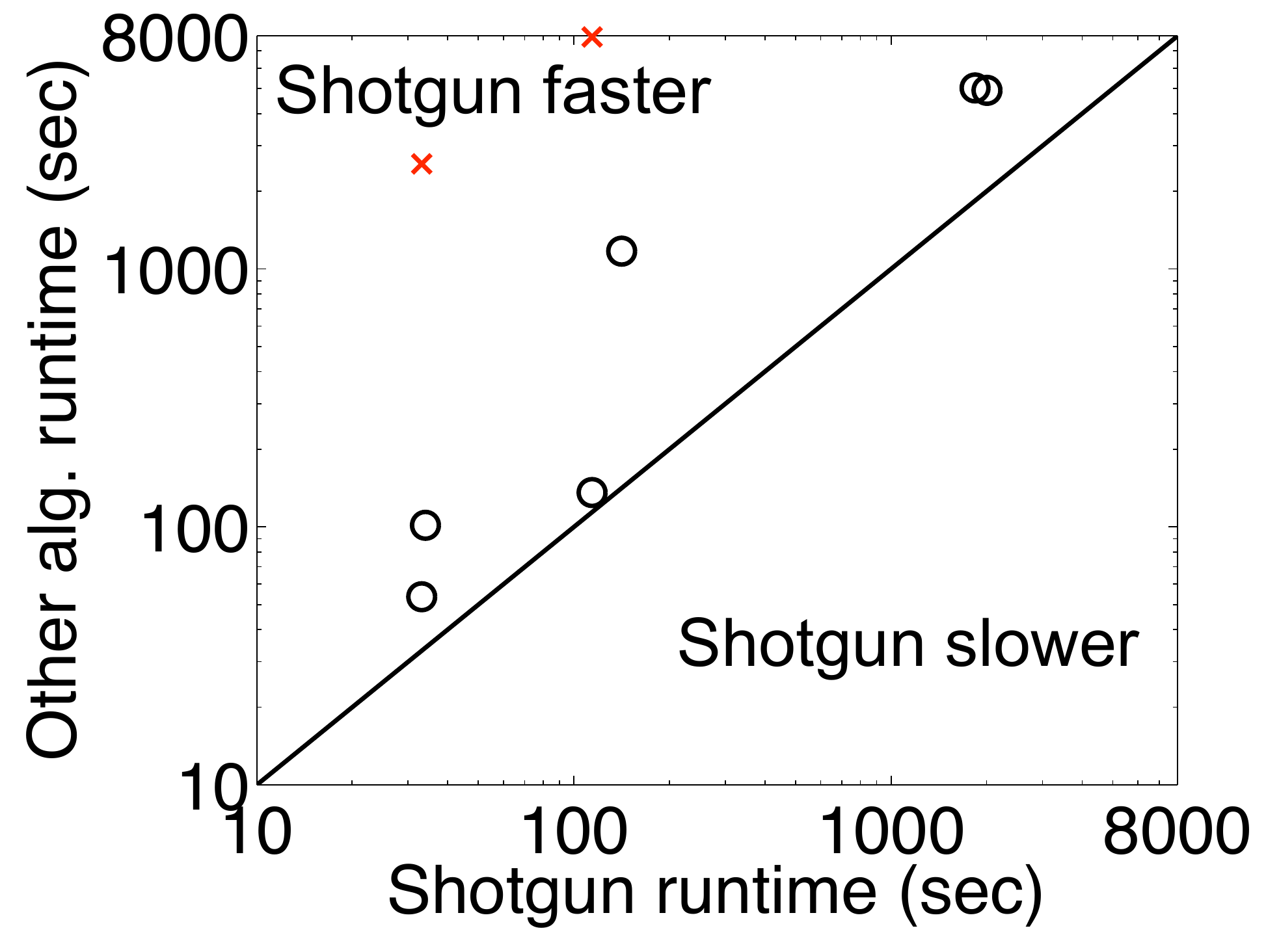}  \\
    \small (a) Sparco & \small (b) Single-Pixel Camera & \small (c) Sparse Compressed Img. & \small (d) Large, Sparse Datasets \normalsize \\
    \small $\p^* \in [3, 17366]$, avg 2987 & \small $\p^* = 3$ & \small $\p^* \in [2865, 11779]$, avg 7688 & \small $\p^* \in [214, 2072]$, avg 1143 \normalsize \\
    \multicolumn{4}{c}{\includegraphics[width=.65\textwidth]{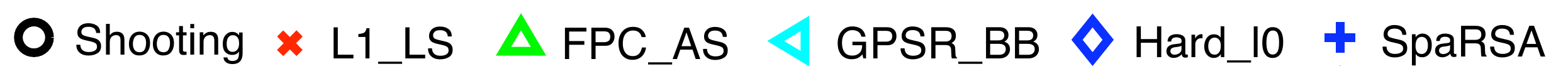}}
  \end{tabular}
  \caption{ Runtime comparison of algorithms for the Lasso on 4 dataset categories.
            Each marker compares an algorithm with {\tt Shotgun} (with $\p=8$) on one dataset (and one $\lambda \in \{0.5, 10\}$).
            Y-axis is that algorithm's running time; X-axis is {\tt Shotgun}'s (P=8) running time on the same problem.
            Markers above the diagonal line indicate that {\tt Shotgun} was faster; markers below the line indicate {\tt Shotgun} was slower.
   \label{fig:lassocomp} }
\end{figure*}

We divide our comparisons into four categories of datasets;
the supplementary material has descriptions.

\begin{changemargin}{-2em}{-2em}
\begin{verse}
\emph{Sparco}: Real-valued datasets of varying sparsity from the Sparco testbed \cite{van2009sparco}. \newline
$n \in [128 29166]$, $d \in [128, 29166]$. \\
\emph{Single-Pixel Camera}: Dense compressed sensing problems from Duarte et al. \yrcite{duarte2008single}. \newline
$n \in [410, 4770]$, $d \in [1024, 16384]$. \\
\emph{Sparse Compressed Imaging}: Similar to Single-Pixel Camera datasets, but with very sparse random $-1/+1$ measurement matrices. Created by us. \newline
$n \in [477, 32768]$, $d \in [954, 65536]$. \\
\emph{Large, Sparse Datasets}: Very large and sparse problems, including predicting stock volatility from text in financial reports \cite{kogan2009predicting}. \newline
$n \in [30465, 209432]$, $d \in [209432, 5845762]$.
\end{verse}
\end{changemargin}

We ran each algorithm on each dataset with regularization $\lambda = 0.5$ and $10$.
\figref{fig:lassocomp} shows runtime results, divided by dataset category.
We omit runs which failed to converge within a reasonable time period.

{\tt Shotgun} (with $\p=8$) consistently performs well, converging faster than other algorithms on most dataset categories.
{\tt Shotgun} does particularly well on the Large, Sparse Datasets category,
for which most algorithms failed to converge anywhere near the ranges plotted in \figref{fig:lassocomp}.
The largest dataset, whose features are occurrences of bigrams in financial reports \cite{kogan2009predicting}, has 5 million features and 30K
samples. On this dataset, {\tt Shooting} converges but requires $\sim 4900$ seconds, while {\tt Shotgun} takes $<2000$ seconds.

On the Single-Pixel Camera datasets, {\tt Shotgun} ($\p=8$) is slower than {\tt Shooting}.
In fact, it is surprising that {\tt Shotgun} converges at all with $\p = 8$,
for the plotted datasets all have $\p^* = 3$.
\figref{fig:shotgun:ptheory} shows {\tt Shotgun} with $\p > 4$ diverging for the {\tt Ball64\_singlepixcam} dataset;
however, after the practical adjustments to {\tt Shotgun} used to produce \figref{fig:lassocomp},
{\tt Shotgun} converges with $\p = 8$.


Among the other solvers, {\tt L1\_LS} is the most robust and even solves some of the Large, Sparse Datasets.

It is difficult to compare optimization algorithms and their implementations.
Algorithms' termination criteria differ;
e.g., primal-dual methods such as {\tt L1\_LS} monitor the duality gap,
while {\tt Shotgun} monitors the change in $\x$.
{\tt Shooting} and {\tt Shotgun} were written in C++, which is generally fast;
the other algorithms were in Matlab, which handles loops slowly but linear algebra quickly.
Therefore, we emphasize major trends:
{\tt Shotgun} robustly handles a range of problems;
\thmref{thm:shotgun} helps explain its speedups;
and {\tt Shotgun} generally outperforms published solvers for the Lasso.

%
%

\subsection{Sparse Logistic Regression}

For logistic regression, we focus on comparing Shotgun with Stochastic Gradient Descent (SGD) variants.
SGD methods are of particular interest to us since they are often considered to be very efficient, especially for learning with many samples;
they often have convergence bounds independent of the number of samples.

For a large-scale comparison of various algorithms for sparse logistic regression,
we refer the reader to the recent survey by Yuan et al. \yrcite{LibLinear}.
On {\tt L1\_logreg} \cite{LogBoyd} and CDN \cite{LibLinear}, our results qualitatively matched their survey.
Yuan et al. \yrcite{LibLinear} do not explore SGD empirically.

\subsubsection{Implementation: {\tt Shotgun~CDN}}

As Yuan et al. \yrcite{LibLinear} show empirically, their Coordinate Descent Newton (CDN) method
is often orders of magnitude faster than the basic Shooting algorithm (\algref{alg:shooting}) for sparse logistic regression.
Like Shooting, CDN does coordinate descent, but instead of using a fixed step size,
it uses a backtracking line search starting at a quadratic approximation of the objective.

Although our analysis uses the fixed step size in \eqref{eqn:shooting:update},
we modified Shooting and Shotgun to use line searches as in CDN.
We refer to CDN as {\tt Shooting CDN}, and we refer to parallel CDN as {\tt Shotgun~CDN}.

{\tt Shooting CDN} and  {\tt Shotgun~CDN} maintain an \emph{active set} of weights which are allowed to become non-zero; 
this scheme speeds up optimization, though it can limit parallelism by shrinking $d$.

\subsubsection{Other Algorithms}

{\tt SGD} iteratively updates $\x$ in a gradient direction estimated with one sample and scaled by a learning rate.
We implemented {\tt SGD} in C++ following, e.g., Zinkevich et al. \yrcite{SmolaGD}.
We used lazy shrinkage updates \cite{Langford+al:truncgrad} to make use of sparsity in $\A$.
Choosing learning rates for SGD can be challenging.
In our tests, constant rates led to faster convergence than decaying rates (decaying as $1/\sqrt{T}$).
For each test, we tried 14 exponentially increasing rates in $[10^{-4}, 1]$ (in parallel) and chose the rate giving the best training objective.
We did not use a sparsifying step for {\tt SGD}.

{\tt SMIDAS} \cite{Shalev-Shwartz+Tewari:ICML09} uses stochastic mirror descent but truncates gradients to sparsify $\x$.
We tested their published C++ implementation.

{\tt Parallel~SGD} refers to Zinkevich et al. \yrcite{SmolaGD}'s work,
which runs {\tt SGD} in parallel on different subsamples of the data and averages the solutions $\x$.
We tested this method since it is one of the few existing methods for parallel regression,
but we note that Zinkevich et al. \yrcite{SmolaGD} did not address $\ellone$ regularization in their analysis.
We averaged over 8 instances of {\tt SGD}.

\subsubsection{Results}

\begin{figure}[t]
  \begin{tabular}{c|c}
    \includegraphics[width=0.23\textwidth,clip]{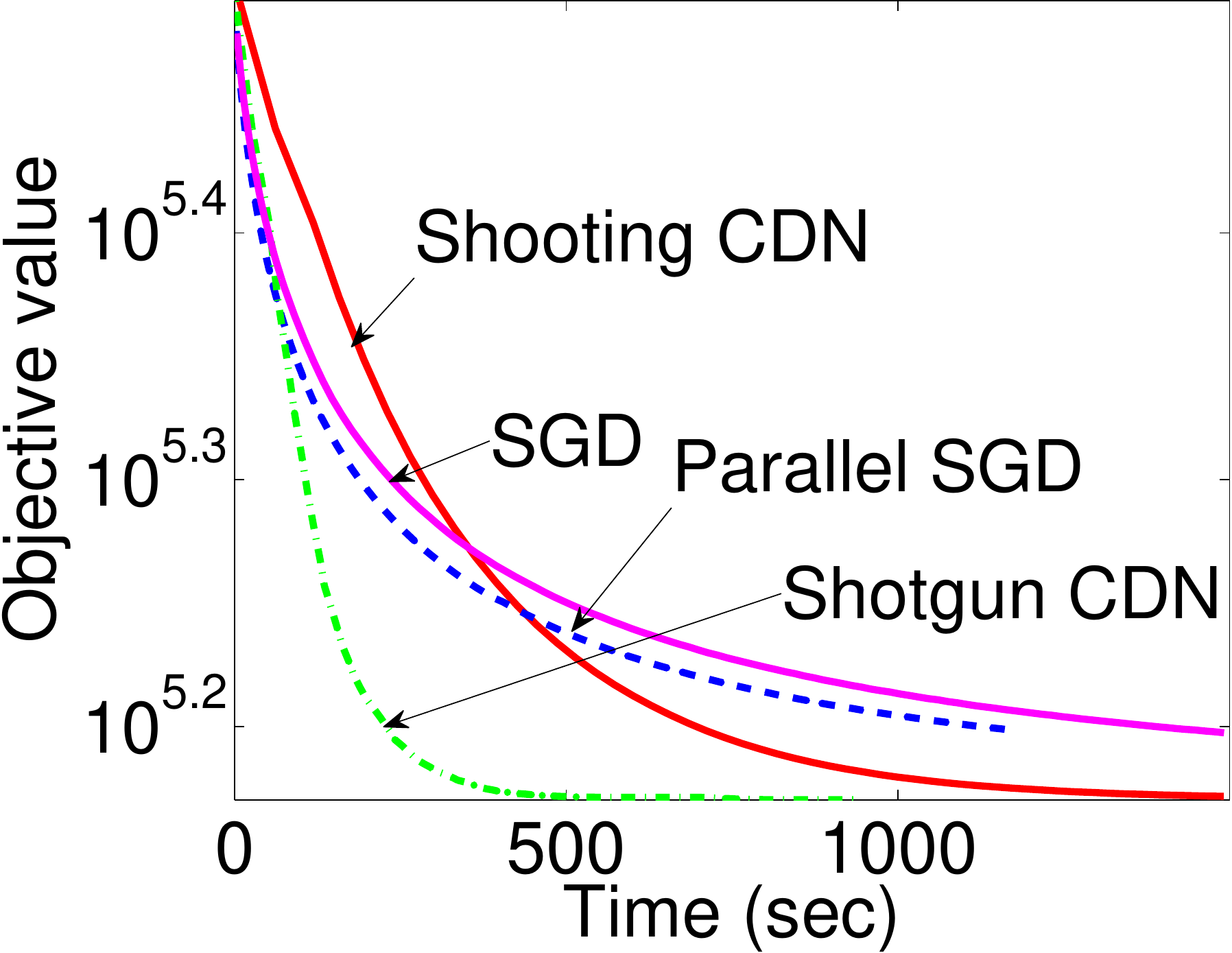} & \includegraphics[width=0.23\textwidth]{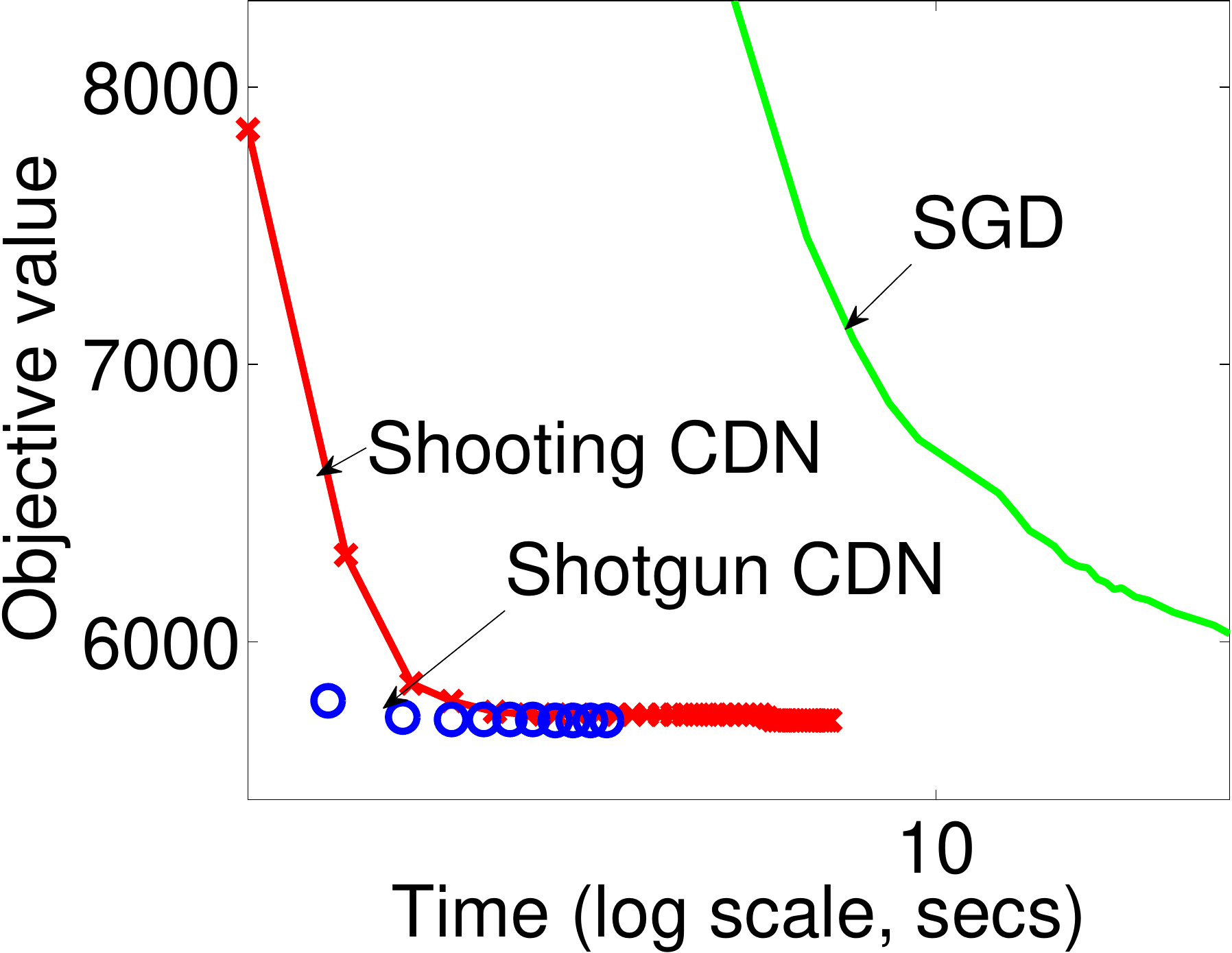} \\
    \includegraphics[width=0.23\textwidth,clip]{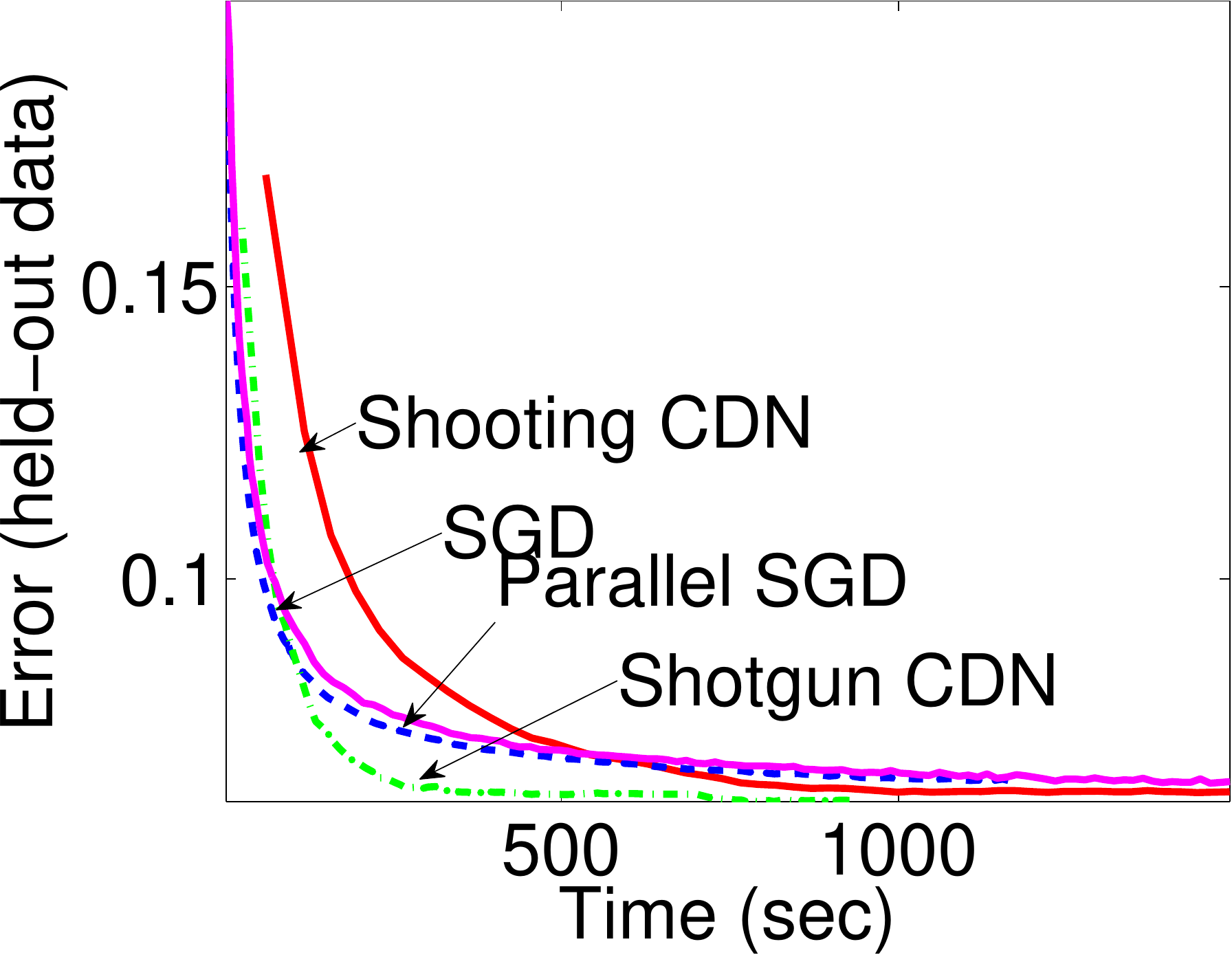} & \includegraphics[width=0.23\textwidth]{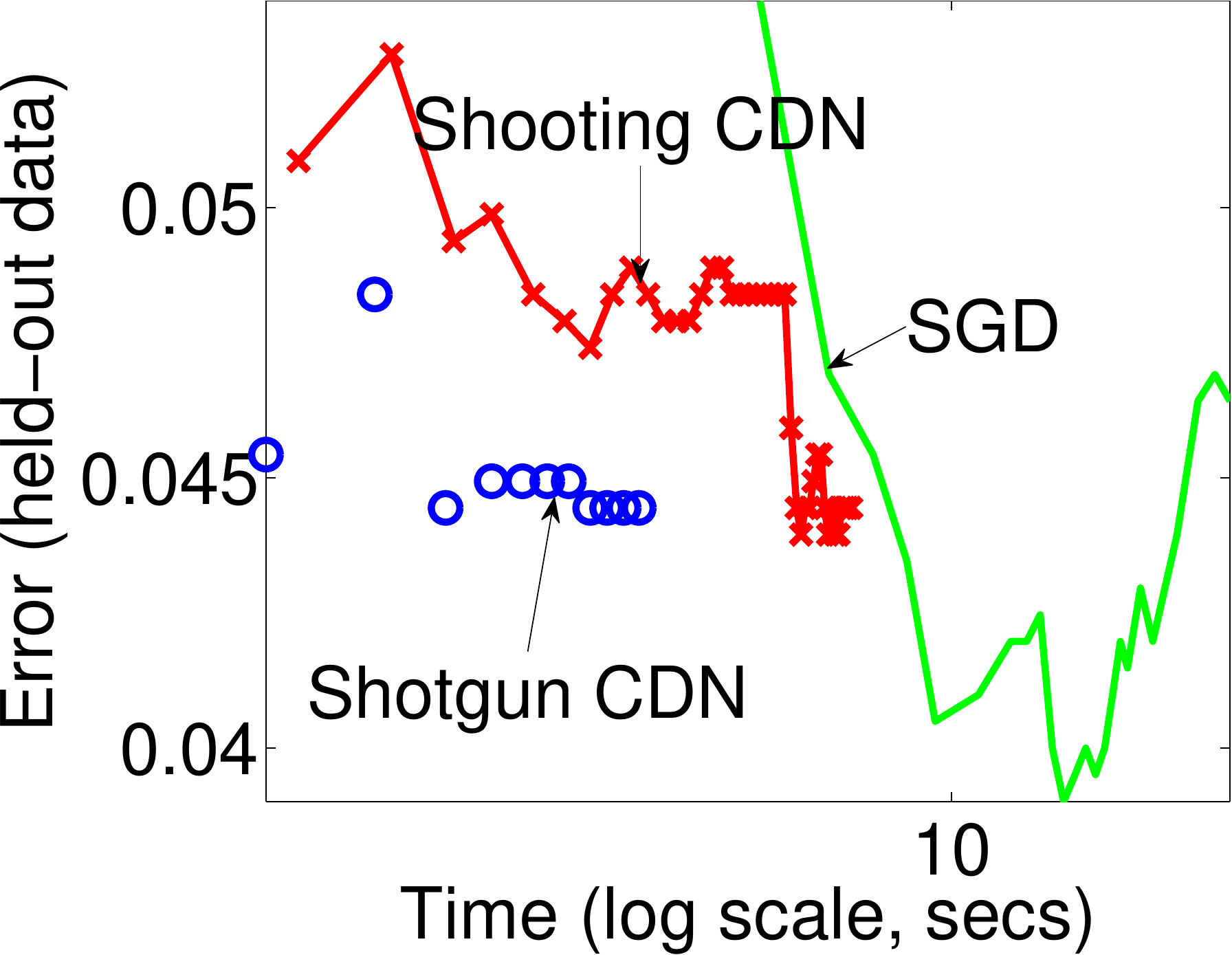} \\
    \small {\tt zeta} with $\lambda=1$ & \small {\tt rcv1} with $\lambda=1$ \\
    \small $d=2000, n=500,000$ & \small $d=44504, n=18217$ \normalsize
  \end{tabular}
  \caption{ Sparse logistic regression on 2 datasets.
   Top plots trace training objectives over time; bottom plots trace classification error rates on held-out data ($10\%$).
   On {\tt zeta} ($n \gg d$), {\tt SGD} converges faster initially, but {\tt Shotgun~CDN} (P=8) overtakes it.
   On {\tt rcv1} ($d > n$), {\tt Shotgun~CDN} converges much faster than {\tt SGD} (note the log scale); {\tt Parallel~SGD} (P=8) is hidden by {\tt SGD}.
   \label{fig:shotgun:sgd} }
\end{figure}

\figref{fig:shotgun:sgd} plots training objectives and test accuracy (on a held-out $10\%$ of the data) for two large datasets.

The {\tt zeta} dataset \footnote{The {\tt zeta} dataset is from the
Pascal Large Scale Learning Challenge: \url{http://www.mlbench.org/instructions/}} illustrates the regime with $n \gg d$.
It contains 500K samples with 2000 features and is fully dense (in $\A$). SGD performs well and is fairly competitive with {\tt Shotgun~CDN} (with $\p=8$).

The {\tt rcv1} dataset \footnote{Our version of the {\tt rcv1} dataset is from the LIBSVM repository \cite{CC01a}.} \cite{rcv1}
illustrates the high-dimensional regime ($d > n$).
It has about twice as many features (44504) as samples (18217), with $17\%$ non-zeros in $\A$.
{\tt Shotgun~CDN} ($\p=8$) was much faster than {\tt SGD}, especially in terms of the objective.
{\tt Parallel~SGD} performed almost identically to {\tt SGD}.

Though convergence bounds for {\tt SMIDAS} are comparable to those for {\tt SGD},
{\tt SMIDAS} iterations take much longer due to the mirror descent updates.
To execute 10M updates on the {\tt zeta} dataset, {\tt SGD} took 728 seconds, while {\tt SMIDAS} took over 8500 seconds.

These results highlight how SGD is orthogonal to Shotgun:
SGD can cope with large $n$, and Shotgun can cope with large $d$.
A hybrid algorithm might be scalable in both $n$ and $d$ and, perhaps, be parallelized over both samples and features.

\subsection{Self-Speedup of Shotgun \label{sec:shotgun:speedups}}

To study the self-speedup of {\tt Shotgun Lasso} and {\tt Shotgun~CDN},
we ran both solvers on our datasets with varying $\lambda$, using varying $\p$ (number of parallel updates = number of cores).
We recorded the running time as the first time when an algorithm came within $0.5\%$ of the optimal objective, as computed by {\tt Shooting}.

\begin{figure*}[t]
  \begin{tabular}{cccc}
    \includegraphics[width=0.23\textwidth]{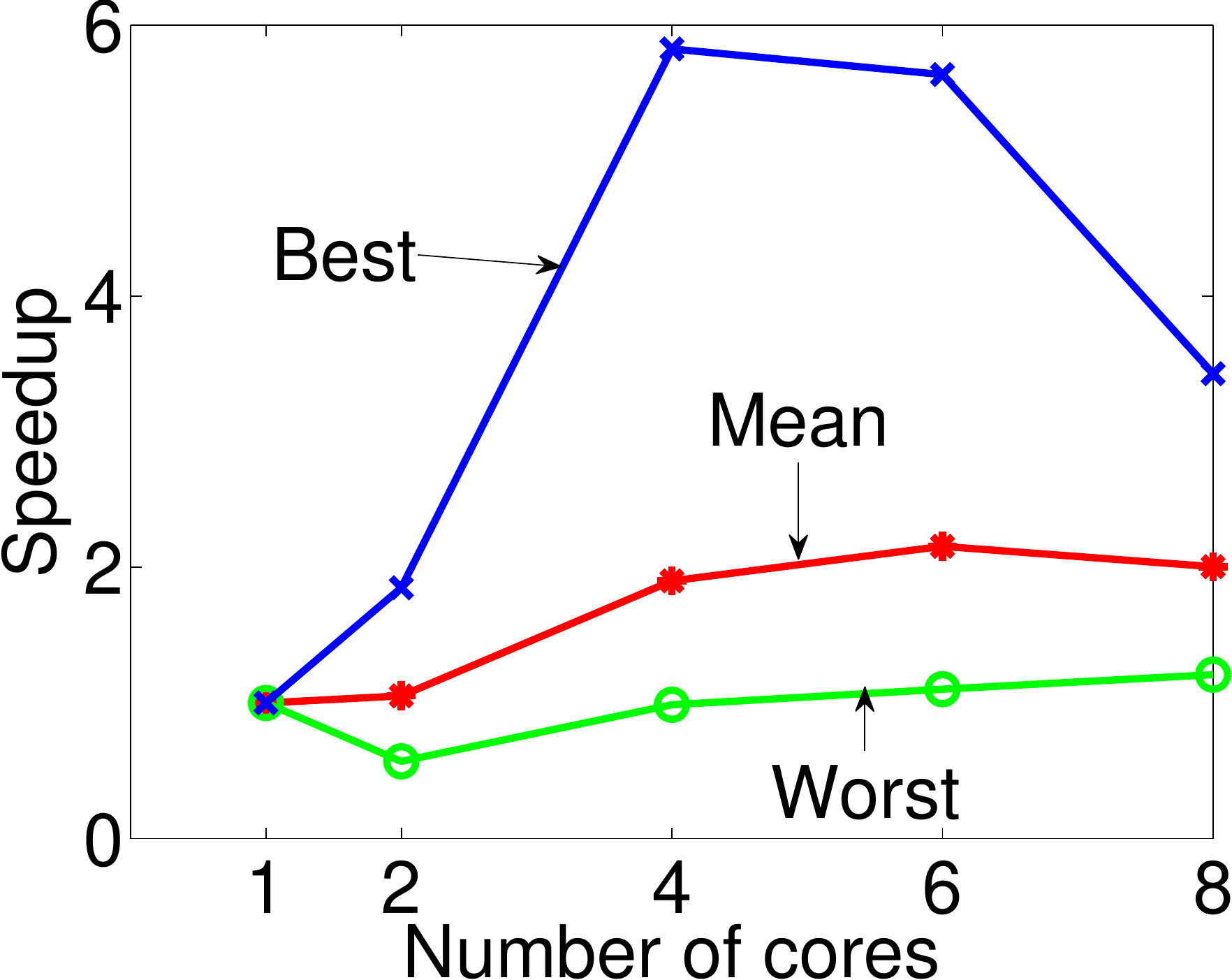} &  \includegraphics[width=0.23\textwidth]{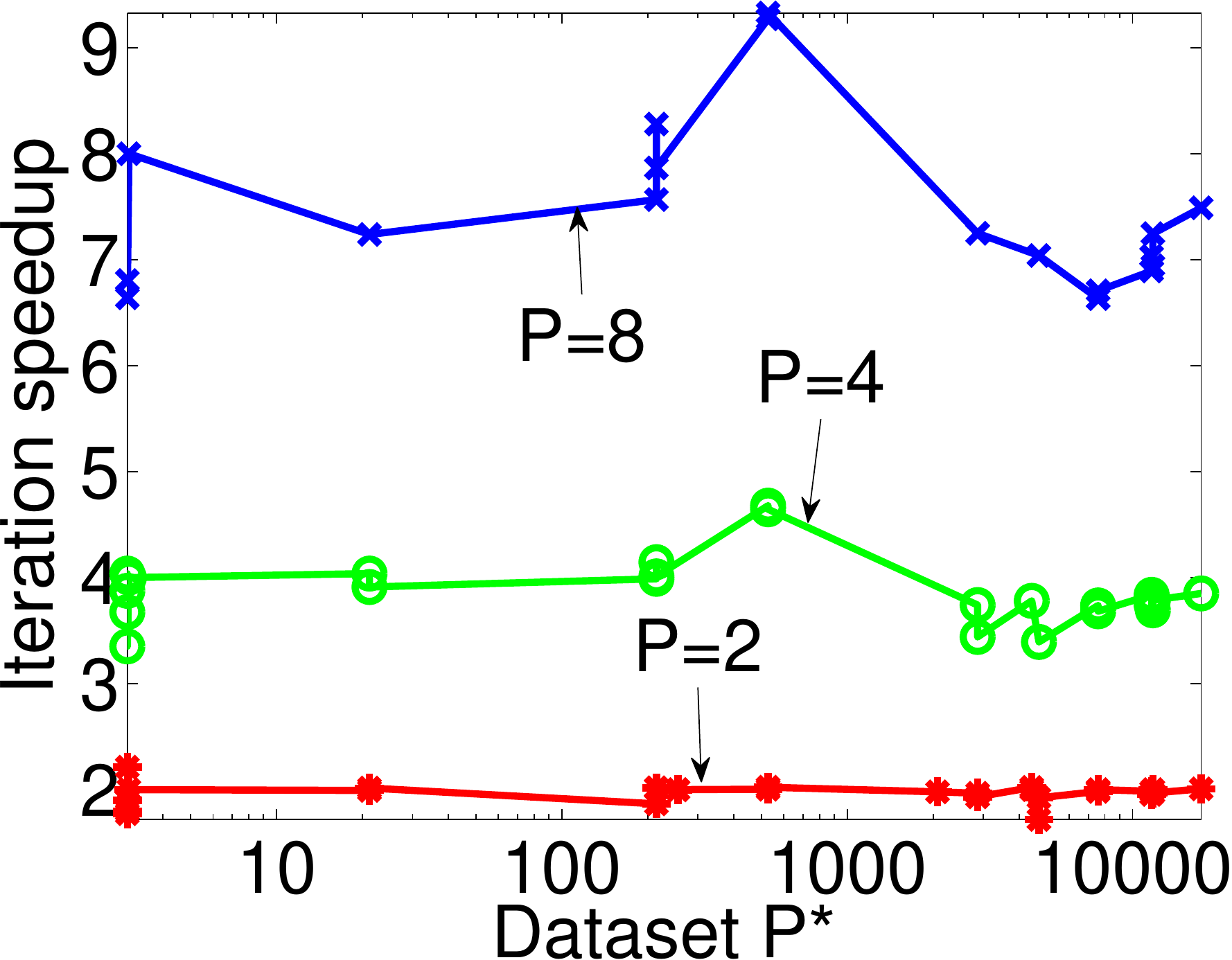}  &  
    \includegraphics[width=0.23\textwidth]{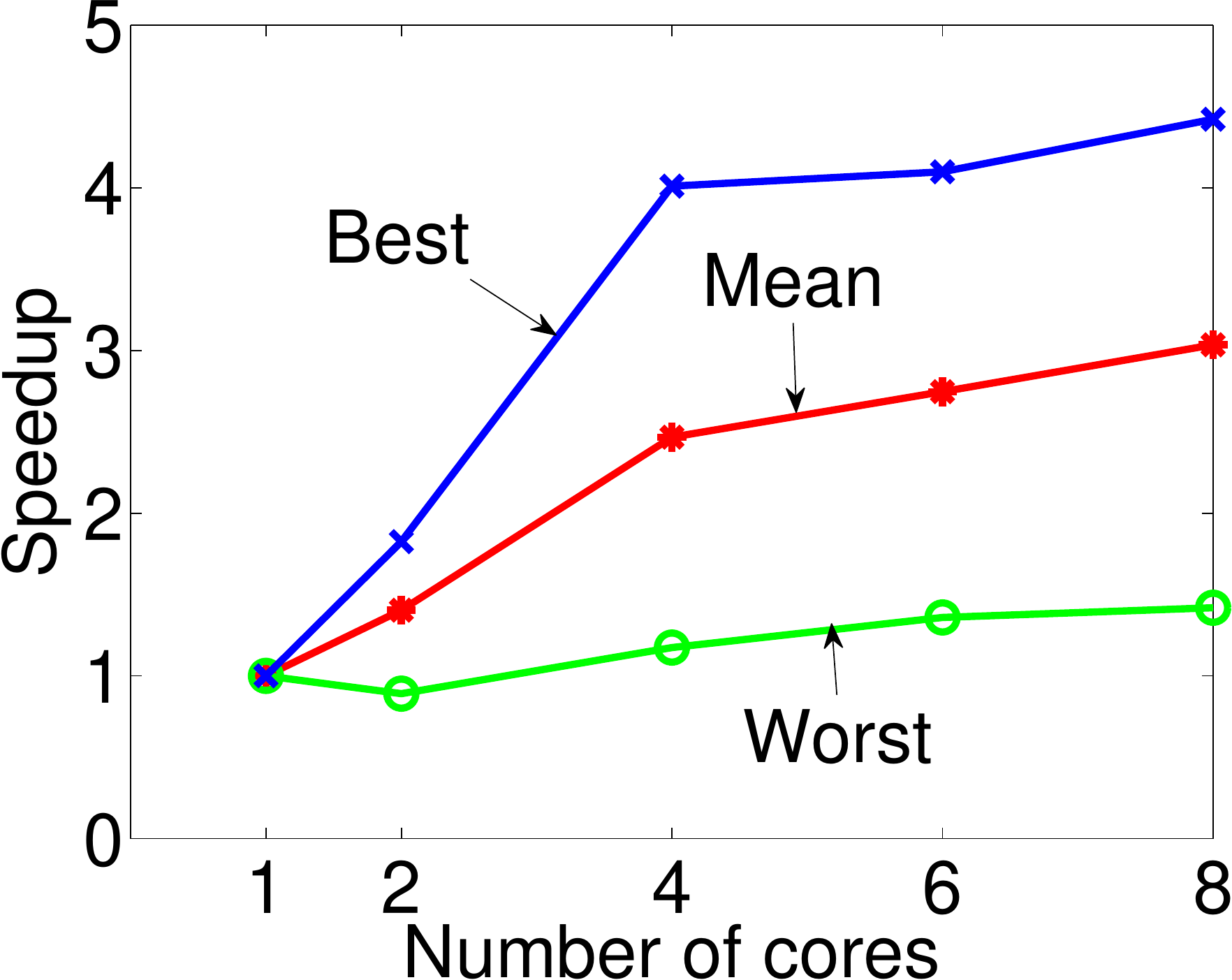} & \includegraphics[width=0.23\textwidth]{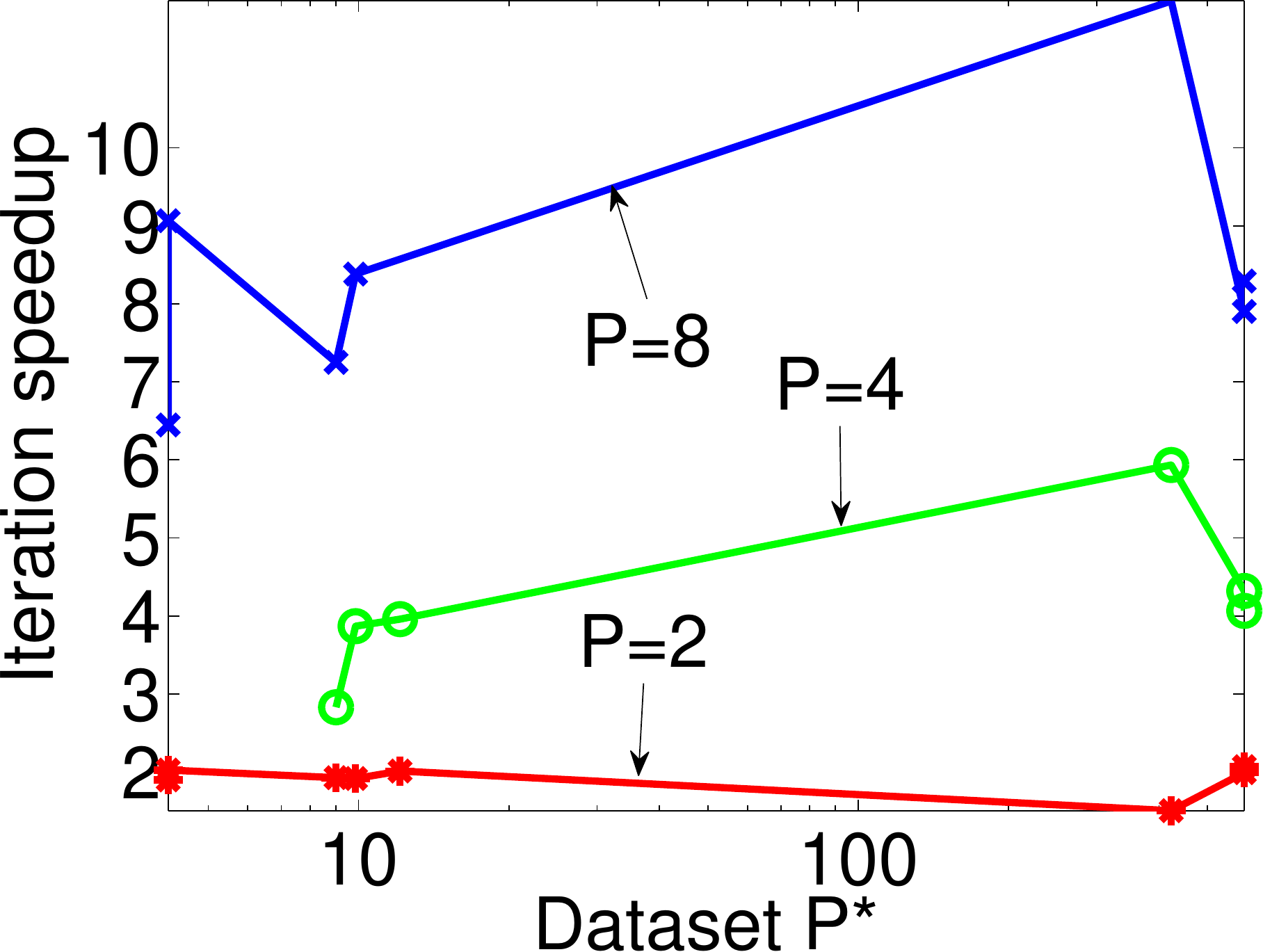} \\
    \small (a) {\tt Shotgun Lasso} & \small (b) {\tt Shotgun Lasso} & \small (c) {\tt Shotgun~CDN} & \small (d) {\tt Shotgun~CDN} \\
    \small runtime speedup & \small iterations & \small runtime speedup & \small iterations \normalsize
  \end{tabular}
  \caption{ \textbf{(a,c)} Runtime speedup in time for {\tt Shotgun Lasso} and {\tt Shotgun~CDN} (sparse logistic regression). \textbf{(b,d)} Speedup in iterations until convergence as a function of $\p^*$.  Both Shotgun instances exhibit almost linear speedups w.r.t. iterations.
      \label{fig:shotgun:speedup} }
\end{figure*}

\figref{fig:shotgun:speedup} shows results for both speedup (in time) and speedup in iterations until convergence.
The speedups in iterations match \thmref{thm:shotgun} quite closely.
However, relative speedups in iterations (about $8\times$) are not matched by speedups in runtime (about $2\times$ to $4\times$).

We thus discovered that speedups in time were limited by low-level technical issues.
To understand the limiting factors, we analyzed various Shotgun-like algorithms to find bottlenecks.\footnote{See the supplement for the scalability analysis details.}
We found we were hitting the \emph{memory wall} \cite{wulf1995hitting};
memory bus bandwidth and latency proved to be the most limiting factors.
Each weight update requires an atomic update to the shared $\A\x$ vector,
and the ratio of memory accesses to floating point operations is only $O(1)$.
Data accesses have no temporal locality since each weight update uses a different column of $\A$.
We further validated these conclusions by monitoring CPU counters.

%
%
%
%
%
%

\section{Discussion}

We introduced the Shotgun, a simple parallel algorithm for $\ellone$-regularized optimization.
Our convergence results for Shotgun are the first such results for parallel coordinate descent with $\ellone$ regularization.
Our bounds predict linear speedups, up to an interpretable, problem-dependent limit.
In experiments, these predictions matched empirical behavior.

Extensive comparisons showed that Shotgun outperforms state-of-the-art $\ellone$ solvers on many datasets.
We believe that, currently, Shotgun is one of the most efficient and scalable solvers for $\ellone$-regularized problems.

The most exciting extension to this work might be the hybrid of SGD and Shotgun discussed in \secref{sec:shotgun:speedups}.

\textbf{Code, Data, and Benchmark Results}: Available at
\textbf{\texttt{http://www.select.cs.cmu.edu/projects}}

\textbf{Acknowledgments} \\
Thanks to John Langford, Guy Blelloch, Joseph Gonzalez, Yucheng Low
and our reviewers for feedback.
Funded by \small NSF IIS-0803333, NSF CNS-0721591, ARO MURI W911NF0710287,
ARO MURI W911NF0810242. \normalsize

%


{
\small
\bibliography{references}
\bibliographystyle{icml2011}
}

\end{document}